\documentclass{article}

\usepackage{graphicx} % more modern
\usepackage{subfigure} 

\usepackage{natbib}

\usepackage{algorithm}
\usepackage{algorithmic}

\usepackage{hyperref}

\usepackage[accepted]{icml2012}

\usepackage{amsmath}
\usepackage{amssymb}

\newcommand{\diff}[2]{\frac{\partial #1}{\partial #2}}
\newcommand{\T}{\top}
\newcommand{\R}{\mathbb{R}}

\icmltitlerunning{Poisoning Attacks against SVMs}

\begin{document} 

\twocolumn[
\icmltitle{Poisoning Attacks against Support Vector Machines}

\icmlauthor{Battista Biggio}{battista.biggio@diee.unica.it}
\icmladdress{Department of Electrical and Electronic Engineering,
  University of Cagliari, Piazza d'Armi, 09123 Cagliari, Italy}
\icmlauthor{Blaine Nelson}{blaine.nelson@wsii.uni-tuebingen.de}
\icmlauthor{Pavel Laskov}{pavel.laskov@uni-tuebingen.de}
\icmladdress{Wilhelm Schickard Institute for Computer Science,
  University of T\"ubingen, Sand 1, 72076 T\"ubingen, Germany}

% You may provide any keywords that you 
% find helpful for describing your paper; these are used to populate 
% the "keywords" metadata in the PDF but will not be shown in the document
\icmlkeywords{}

\vskip 0.3in
]

\begin{abstract} 
  We investigate a family of poisoning attacks against Support Vector
  Machines (SVM). Such attacks inject specially crafted training data
  that increases the SVM's test error.  Central to the motivation for
  these attacks is the fact that most learning algorithms assume that
  their training data comes from a natural or well-behaved
  distribution.  However, this assumption does not generally hold in
  security-sensitive settings. As we demonstrate, an intelligent
  adversary can, to some extent, predict the change of the SVM's
  decision function due to malicious input and use this ability to
  construct malicious data.

  The proposed attack uses a gradient ascent strategy in which the
  gradient is computed based on properties of the SVM's optimal
  solution.  This method can be kernelized and enables
  the attack to be constructed in the \emph{input space} even for
  non-linear kernels. We experimentally demonstrate that our gradient
  ascent procedure reliably identifies good local maxima of the
  non-convex validation error surface, which significantly increases
  the classifier's test error.
\end{abstract} 

\section{Introduction}

% General motivation for attacking Machine Learning algorithms and generally for exploring their security properties.

Machine learning techniques are rapidly emerging as a vital tool in a
variety of networking and large-scale system applications because they
can infer hidden patterns in large complicated datasets, adapt to new
behaviors, and provide statistical soundness to decision-making
processes. Application developers thus can employ learning to help
solve so-called \emph{big-data problems} and these include a number of
security-related problems particularly focusing on identifying
malicious or irregular behavior.  In fact, learning approaches have
already been used or proposed as solutions to a number of such
security-sensitive tasks including spam, worm, intrusion and fraud
detection~\cite{meyer-whateley-2004-spambayes,biggio-IJMLC10,Stolfo-Hershkop-MMM-ACNS-2003,Forrest-Hofmeyr-SP-1996,Bolton-Hand-JSS-2002,CovKruVig10,RieKruDew10,CurLiv11,LasSrn11}.
Unfortunately, in these domains, data is generally not only
non-stationary but may also have an adversarial component, and the
flexibility afforded by learning techniques can be exploited by
an adversary to achieve his goals. For instance, in spam-detection,
adversaries regularly adapt their approaches based on the popular spam
detectors, and generally a clever adversary will change his behavior
either to evade or mislead learning.

% Setting for this paper: Attacking SVMs and the particular type of attack we present.

In response to the threat of adversarial data manipulation, several
proposed learning methods explicitly account for
certain types of corrupted data
\cite{GloRow06,TeoGloRowSmo08,Bruckner-Scheffer-NIPS-2009,DekSha10}.
Attacks against learning algorithms can be classified, among other
categories \citep[c.f.][]{Barreno-Nelson-ML-2010}, into
\emph{causative} (manipulation of training data) and
\emph{exploratory} (exploitation of the classifier). \emph{Poisoning}
refers to a causative attack in which specially crafted attack points
are injected into the training data. This attack is especially
important from the practical point of view, as an attacker usually
cannot directly access an existing training database but may
\emph{provide} new training data; \emph{e.g.}, web-based repositories and
honeypots often collect malware examples for training, which provides
an opportunity for the adversary to poison the training data. Poisoning
attacks have been previously studied only for simple
anomaly detection methods
\cite{Barreno-Nelson-ASIACCS-2006,Rubinstein-Nelson-IMC-2009,Kloft-Laskov-AIStats-2010}.

In this paper, we examine a family of poisoning attacks against
Support Vector Machines (SVM). Following the general security analysis methodology for machine learning, we assume that the attacker
knows the learning algorithm and can draw data from the underlying
data distribution. Further, we assume that our attacker knows the
training data used by the learner; generally, an unrealistic
assumption, but in real-world settings, an attacker could instead use
a surrogate training set drawn from the same distribution
\citep[\emph{e.g.},][]{Nelson-Barreno-LEET-2008} and our approach yields a
worst-case analysis of the attacker's capabilities.  Under these
assumptions, we present a method that an attacker can use to
construct a data point that significantly decreases the SVM's classification
accuracy.

The proposed method is based on the properties of the optimal solution
of the SVM training problem. As was first shown in an incremental
learning technique \cite{CauPog01}, this solution depends smoothly on
the parameters of the respective quadratic programming problem and on
the geometry of the data points.  Hence, an attacker can manipulate
the optimal SVM solution by inserting specially crafted attack points.
We demonstrate that finding such an attack point can be formulated as
optimization with respect to a performance measure, subject to the
condition that an optimal solution of the SVM training problem is
retained. Although the test error surface is generally nonconvex, the
gradient ascent procedure used in our method reliably identifies good
local maxima of the test error surface.

The proposed method only depends on the gradients of the dot products
between points in the input space, and hence can be \emph{kernelized}.
This contrasts previous work involving construction of special attack
points
\citep[\emph{e.g.},][]{Bruckner-Scheffer-NIPS-2009,Kloft-Laskov-AIStats-2010}
in which attacks could only be constructed in the feature space for
the nonlinear case. The latter is a strong disadvantage for the
attacker, since he must construct data in the input space and has no
practical means to access the feature space.  Hence, the proposed
method breaks new ground in \emph{optimizing} the impact of
data-driven attacks against kernel-based learning algorithms and
emphasizes the need to consider resistance against adversarial
training data as an important factor in the design of learning
algorithms.

\section{Poisoning attack on SVM}

We assume the SVM has been trained on a data set $\mathcal
D_{\text{tr}} = \{ x_{i}, y_{i} \}_{i=1}^{n}$, $x_i \in
\R^d$. Following the standard notation, $K$ denotes the matrix of
kernel values between two sets of points, $Q = yy^\T \odot K$ denotes
the label-annotated version of $K$, and $\alpha$ denotes the SVM's dual
variables corresponding to each training point.  Depending on the value
of $\alpha_i$, the training points are referred to as margin support
vectors ($0 < \alpha_i < C$, set $\mathcal{S}$), error support vectors
($\alpha_i = C$, set $\mathcal{E}$) and reserve points ($\alpha_i =
0$, set $\mathcal{R}$). In the sequel, the lower-case letters $s,e,r$ are used to index the
corresponding parts of vectors or matrices; \emph{e.g.}, $Q_{ss}$ denotes the
margin support vector submatrix of $Q$.

\subsection{Main derivation}
\label{sec:derivation}

For a poisoning attack, the attacker's goal is to find a point
$(x_c,y_c)$, whose addition to $\mathcal D_{\text{tr}}$ maximally
decreases the SVM's classification accuracy. The choice of the attack
point's label, $y_c$, is arbitrary but fixed. We refer to the class of
this chosen label as \emph{attacking} class and the other as the
\emph{attacked} class.

The attacker proceeds
by drawing a validation data set $\mathcal D_{\text{val}} = \{
x_{k}, y_{k} \}_{k=1}^{m}$ and maximizing the hinge loss incurred on
$\mathcal D_{\text{val}}$ by the SVM trained on $\mathcal
D_{\text{tr}} \cup (x_c, y_c)$:
\begin{equation}
\label{eq:L_def}
\max_{x_c} L(x_c) = \sum_{k=1}^{m} (1 - y_k f_{x_c}(x_k))_+ =
\sum_{k=1}^{m} (-g_k)_+ \enspace 
\end{equation}
In this section, we assume the role of the attacker and develop a
method for optimizing $x_c$ with this objective.

First, we explicitly account for all terms in the margin conditions
$g_k$ that are affected by $x_c$:
\begin{align}
    g_k &= \sum_{j} Q_{kj} \alpha_j + y_k b - 1 \label{eq:gk_def}\\
    &= \sum_{j \neq c} Q_{kj} \alpha_j (x_c) +
      Q_{kc}(x_c) \alpha_c (x_c) + y_k b(x_c) - 1 \; .\notag
\end{align}

%BB: Blaine, please check below.
It is not difficult to see from the above equations that $L(x_{c})$ is a non-convex objective function.
Thus, we exploit a gradient ascent technique to iteratively optimize it.
We assume that an initial location of the attack point $x_c^{(0)}$ has been chosen.
Our goal is to update the attack point as $ x_c^{(p)} =  x_c^{(p-1)} + t u$ where $p$ is the current iteration, $u$ is a norm-1 vector representing the attack direction, and $t$ is the step size.
Clearly, to maximize our objective, the attack direction $u$ aligns to the gradient of $L$ with respect to $u$, which has to be computed at each iteration.  

Although the hinge loss is not everywhere differentiable,
this can be overcome by only considering point
indices $k$ with non-zero contributions to
$L$; \emph{i.e.}, those for which $-g_k > 0$. Contributions of such points to
the gradient of $L$ can be computed by differentiating
Eq.~(\ref{eq:gk_def}) with respect to $u$ using the product rule:

\begin{equation}
  \label{eq:gk_diff}
  \diff{g_k}{u} = Q_{ks} \diff{\alpha}{u} +
  \diff{Q_{kc}}{u} \alpha_c + y_k \diff{b}{u},
\end{equation}
where
\begin{equation*}
  \diff{\alpha}{u} = 
  \begin{bmatrix}
    \diff{\alpha_1}{u_1} & \cdots & \diff{\alpha_1}{u_d} \\
    \vdots & \ddots & \vdots\\
    \diff{\alpha_s}{u_1} & \cdots & \diff{\alpha_s}{u_d}
  \end{bmatrix}, \;\text{simil.}\; \diff{Q_{kc}}{u},\, \diff{b}{u} \enspace .
\end{equation*}

The expressions for the gradient can be further refined using the fact
that the step taken in direction $u$ should maintain the optimal SVM
solution. This can expressed as an adiabatic update condition using
the technique introduced in \cite{CauPog01}. Observe that for the
$i$-th point in the \emph{training} set, the KKT conditions for the
optimal solution of the SVM training problem can be expressed as:
%\begin{equation}
\begin{align}
g_{i} &=   \sum_{j \in \mathcal{D}_{\text{tr}}} Q_{ij} \alpha_{j}+y_{i}b-1
\begin{cases}
> 0; \;  i \in \mathcal R\\
= 0; \;  i \in \mathcal S \\
< 0; \;  i \in \mathcal E
\end{cases} \label{eq:kt-svm-1}\\
h &=  \sum_{j \in \mathcal{D}_{\text{tr}}} y_{j}\alpha_{j}=0 \label{eq:kt-svm-2} \enspace .
\end{align}
%\end{equation}
The equality in condition \eqref{eq:kt-svm-1} and
\eqref{eq:kt-svm-2} implies that an infinitesimal change in the attack
point $x_c$ causes a smooth change in the optimal solution of the
SVM, under the restriction that the composition of the sets
$\mathcal{S}$, $\mathcal{E}$ and $\mathcal{R}$ remain intact. This
equilibrium allows us to predict the \emph{response} of the SVM solution
to the variation of $x_c$, as shown below.

By differentiation of the $x_c$-dependent terms in
Eqs. \eqref{eq:kt-svm-1}--\eqref{eq:kt-svm-2} with respect to each
component $u_l$ ($1 \leq l \leq d$), we obtain, for any $i \in \mathcal S$,
\begin{equation}
  \label{eq:kt_diff}
  \begin{aligned}
  \diff{g}{u_l} &= Q_{ss} \diff{\alpha}{u_l} + \diff{Q_{sc}}{u_l} \alpha_c +
    y_s \diff{b}{u_l} = 0\\
    \diff{h}{u_l} &= y_s^\T \diff{\alpha}{u_l} = 0 \enspace ,
  \end{aligned}
\end{equation}
which can be rewritten as
\begin{equation}
  \label{eq:kt_diff_mat}
  \begin{aligned}
  \begin{bmatrix}
    \diff{b}{u_l} \\ \diff{\alpha}{u_l}
  \end{bmatrix} &=  -
  {\begin{bmatrix}
    0 & y_S^\T \\
    y_s & Q_{ss}
  \end{bmatrix}}^{-1}
  \begin{bmatrix}
    0 \\ \diff{Q_{sc}}{u_l}
  \end{bmatrix}
  \alpha_c %\\
%   &= 
%   -R  
%   \begin{bmatrix}
%     0 \\ \diff{Q_{sc}}{u_l}
%   \end{bmatrix}
%   \alpha_c 
  \enspace .
  \end{aligned}
\end{equation}
The first matrix can be inverted using the
Sherman-Morrison-Woodbury formula~\citep{Lue96}:
\begin{equation}
  \label{eq:smw}
  \begin{aligned}
    \begin{bmatrix}
      0 & y_s^{\T} \\
      y_s & Q_{ss}
    \end{bmatrix}^{-1}
    = \frac{1}{\zeta}
    \begin{bmatrix}
      -1 & \upsilon^\T \\
      \upsilon & \zeta Q_{ss}^{-1} - \upsilon \upsilon^{\T}
    \end{bmatrix}
  \end{aligned}
\end{equation}
where $\upsilon = Q_{ss}^{-1} y_s$ and $\zeta = y_s^\T Q_{ss}^{-1}
y_s$.
Substituting (\ref{eq:smw}) into (\ref{eq:kt_diff_mat}) and observing
that all components of the inverted matrix are independent of $x_c$, we obtain:
\begin{equation}
  \label{eq:adiabat}
  \begin{aligned}
    \diff{\alpha}{u} &= - \frac{1}{\zeta} \alpha_c (\zeta Q_{ss}^{-1} - \upsilon
    \upsilon^\T) \cdot \diff{Q_{sc}}{u} \\
    \diff{b}{u} &= - \frac{1}{\zeta} \alpha_c  \upsilon^\T \cdot \diff{Q_{sc}}{u} \enspace .
  \end{aligned}
\end{equation}
Substituting (\ref{eq:adiabat}) into (\ref{eq:gk_diff}) and further
into (\ref{eq:L_def}), we obtain the desired gradient used for optimizing our attack:
\begin{equation}
  \label{eq:L_grad}
  \diff{L}{u} = \sum_{k=1}^m  \left \{   M_k
  \diff{Q_{sc}}{u} +  \diff{Q_{kc}}{u} \right \} \alpha_{c} , 
\end{equation}
where
\begin{equation*}
  M_k = -\frac{1}{\zeta} ( Q_{ks} (\zeta Q_{ss}^{-1} - \upsilon\upsilon^{T})  + y_{k} \upsilon^{T}).
\end{equation*}

\subsection{Kernelization}

From Eq.~(\ref{eq:L_grad}), we see that the gradient of the objective function at iteration $k$ may depend on the attack point $x_{c}^{(p)}=x_{c}^{(p-1)}+t u$ only through the gradients of the matrix $Q$. In particular, this depends on the chosen kernel. We report below the expressions of these gradients for three common kernels.

%BAT: I would revise these expressions---
\begin{itemize}
\item Linear kernel:
  $$
  %\diff{(x_i \cdot (x_c^0 + t u))}{u} = t x_i
  \diff{K_{ic}}{u} = \diff{(x_i \cdot x_c^{(p)})}{u} = t x_i
  $$
\item Polynomial kernel:
  $$
  %\diff{(x_i \cdot (x_c^0 + t u) + R)^d}{u} = d(x_i \cdot (x_c^0 + t u) + R)^{d-1} t x_i
  \diff{K_{ic}}{u} = \diff{(x_i \cdot x_c^{(p)}+ R)^d}{u} = d(x_i \cdot x_{c}^{(p)} + R)^{d-1} t x_i 
  $$
\item RBF kernel:
  $$
  %\diff{e^{-\frac{\gamma}{2} ||x_i - x_c^0 - t u||^2}}{u} = K(x_i,x_c) \gamma t (x_i - x_c^0 - t u)
  \diff{K_{ic}}{u} = \diff{e^{-\frac{\gamma}{2} ||x_i - x_c||^2}}{u} = K(x_i,x_c^{(p)}) \gamma t (x_i - x_c^{(p)})
  $$
\end{itemize}

The dependence on $x_{c}^{(p)}$ (and, thus, on $u$) in the gradients of non-linear kernels can be avoided by substituting $x_{c}^{(p)}$ with $x_{c}^{(p-1)}$, provided that $t$ is sufficiently small. This approximation enables a straightforward extension of our method to arbitrary kernels.

\subsection{Poisoning Attack Algorithm}

The algorithmic details of the method described in
Section~\ref{sec:derivation} are presented in
Algorithm~\ref{alg:poison}. 

In this algorithm, the attack vector $x_c^{(0)}$ is initialized by cloning
an arbitrary point from the attacked class and flipping its label. In
principle, any point \emph{sufficiently deep} within the attacking
class's margin can be used as a starting point. However, if this point
is too close to the boundary of the attacking class, the iteratively
adjusted attack point may become a reserve point, which halts further
progress.

\begin{algorithm}[tb]
  \caption{Poisoning attack against SVM}
  \label{alg:poison}
  \textbf{Input:} $\mathcal D_{\rm tr}$, the training data; $\mathcal D_{\rm val}$, the validation data; $y_{c}$, the class label of the attack point; $x_{c}^{(0)}$, the initial attack point; $t$, the step size.\\
  \textbf{Output:} $x_{c}$, the final attack point.
  
  \begin{algorithmic}[1]
        \STATE{ $\{ \alpha_{i}, b \} \leftarrow $ learn an SVM on $\mathcal D_{\rm tr}$.}
    \STATE{$k \gets 0$.}
    \REPEAT
            \STATE{Re-compute the SVM solution on $\mathcal D_{\rm tr} \cup \{x_c^{(p)}, y_{c} \}$ using incremental SVM
      \citep[\emph{e.g.},][]{CauPog01}. This step requires $\{ \alpha_{i}, b \}$.}
     \STATE{Compute $\diff{L}{u}$ on $\mathcal{D}_{\rm val}$ according to
      Eq.~\eqref{eq:L_grad}.}
      \STATE{Set $u$ to a unit vector aligned
      with $\diff{L}{u}$.} 
      \STATE{$k \gets k + 1$ and $x_c^{(p)} \gets  x_c^{(p-1)} + t u$}
      \UNTIL{$L\left(x_c^{(p)}\right) - L\left( x_c^{(p-1)}\right) < \epsilon$}
      \STATE{\textbf{return:} $x_c = x_c^{(p)}$}
  \end{algorithmic}
\end{algorithm}

The computation of the gradient of the validation error crucially
depends on the assumption that the structure of the sets
$\mathcal{S}$, $\mathcal{E}$ and $\mathcal{R}$ does not change during
the update. In general, it is difficult to determine the largest step
$t$ along an arbitrary direction $u$, which preserves this structure.
The classical line search strategy used in gradient ascent
methods is not suitable for our case, since the update to the optimal
solution for large steps may be prohibitively expensive. Hence, the
step $t$ is fixed to a small constant value in our algorithm. After
each update of the attack point $x_c^{(p)}$, the optimal solution is %updated
efficiently recomputed from the solution on $\mathcal{D}_{\rm tr}$,
using the incremental SVM machinery \citep[\emph{e.g.},][]{CauPog01}.
%If any structural changes take place along the step $t$, the inverse matrix $R$ in Eq.~\eqref{eq:smw} must be updated with the cost of $O(s^2)$; otherwise the updates are linear in the number of training points.
%BB: I would remove the above part, since R changes at each iteration when xc is a margin SV. So, NOW, we know that we can still mantain R with rank one updates (Best's approach), but this can not be explained here!

The algorithm terminates when the change in the validation error is smaller than a predefined threshold. For kernels including the linear
kernel, the surface of the validation error is unbounded, hence the
algorithm is halted when the attack vector deviates too much
from the training data; \emph{i.e.}, we bound the size of our attack points.

%discussion on termination criteria: looking far behind, not just at the prev step. Improvement: as we keep the svm solution updated during the process, a better function to look at is the validation error (instead of L). Blaine, what about the other criterion on xc not moving anymore?

%further improvements: multiple initializations for multiple runs of alg1, eventually retaining the best one. This should avoid accepting a solution which lies on a poor local maximum (let's look for the global one..)

\section{Experiments}

The experimental evaluation presented in the following sections
demonstrates the behavior of our proposed method on an artificial
two-dimensional dataset and evaluates its effectiveness on the
classical MNIST handwritten digit recognition dataset.

\subsection{Artificial data}

We first consider a two-dimensional data generation model in which each
class follows a Gaussian distribution with mean and covariance
matrices given by $\mu_{-} = [-1.5, 0]$, $\mu_{+} = [1.5, 0]$, $\Sigma_{-}
= \Sigma_{+} = 0.6 I$. The points from the \emph{negative} distribution
are assigned the label $-1$ (shown as red in the subsequent figures)
and otherwise $+1$ (shown as blue).  The training and the validation sets,
$\mathcal D_{\rm tr}$ and $\mathcal D_{\rm val}$ (consisting of $25$
and $500$ points per class, respectively) are randomly drawn from this
distribution.

\begin{figure*}[tb]
\begin{center}
\includegraphics[width=0.32\textwidth]{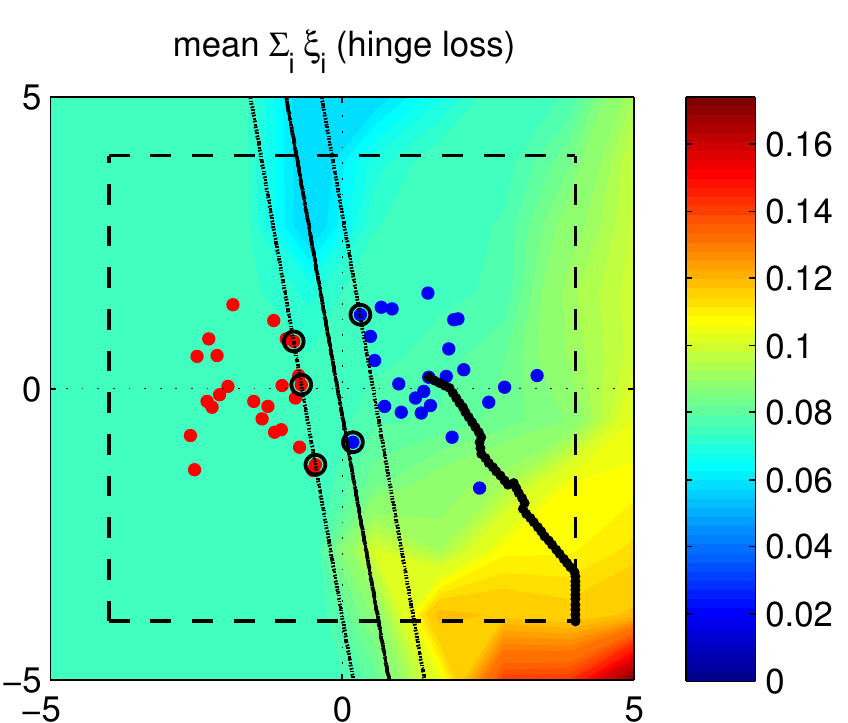}
\includegraphics[width=0.32\textwidth]{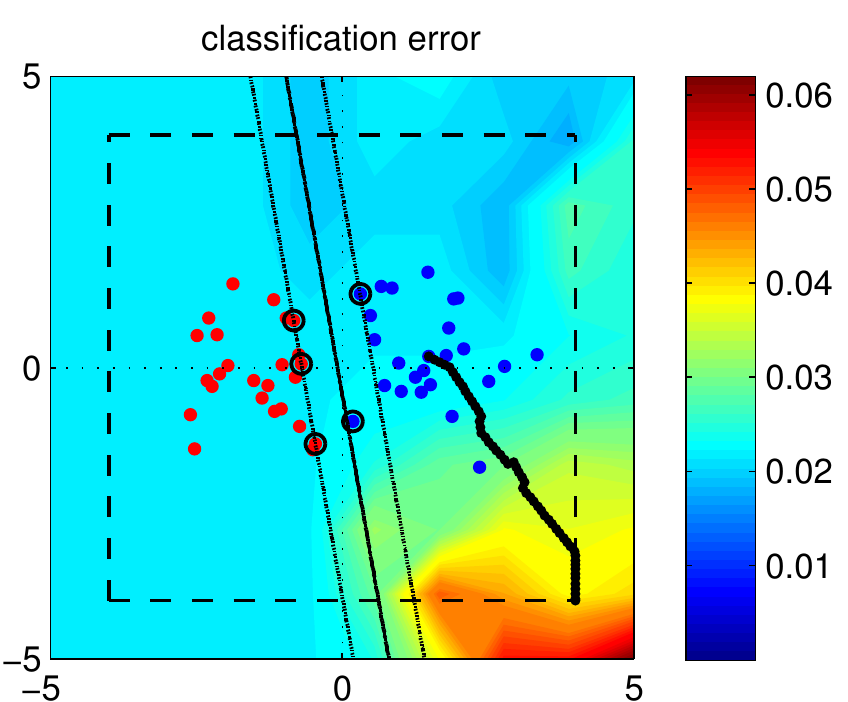}\\
\includegraphics[width=0.32\textwidth]{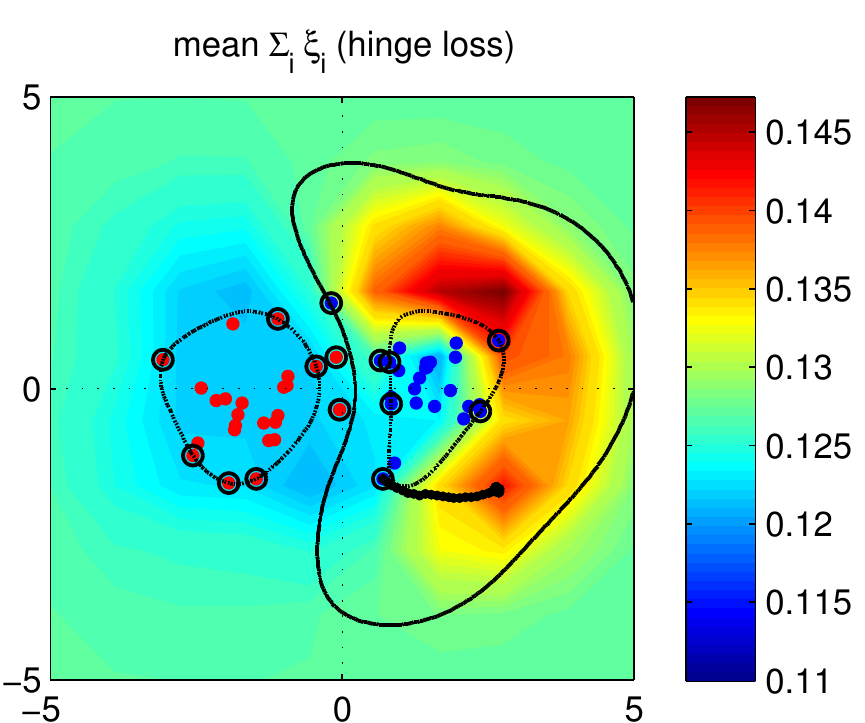}
\includegraphics[width=0.32\textwidth]{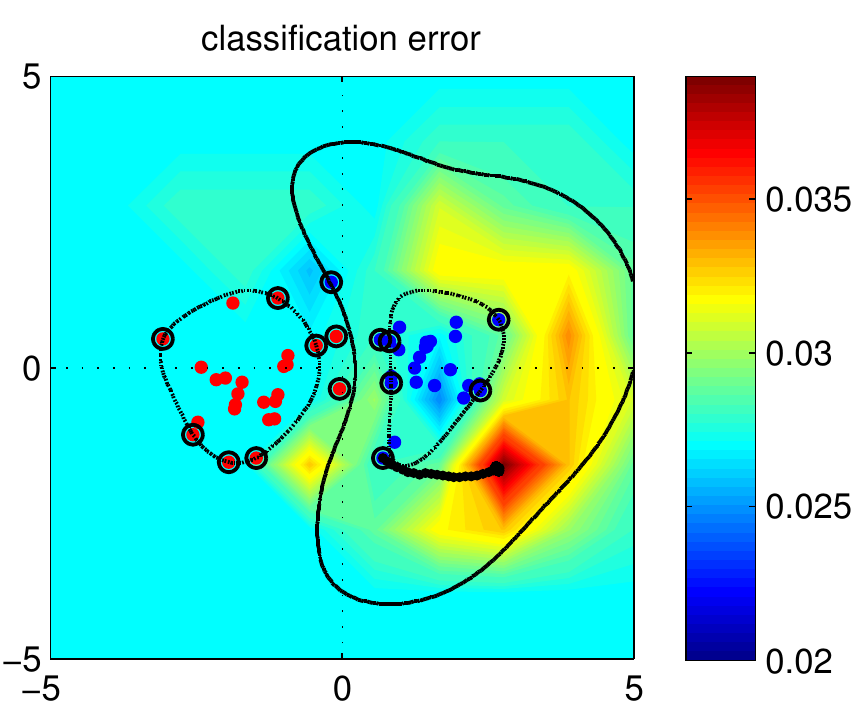}
\caption{Behavior of the gradient-based attack strategy on the
  Gaussian data sets, for the linear (top row) and the RBF kernel
  (bottom row) with $\gamma=0.5$. The regularization parameter $C$ was
  set to $1$ in both cases.  The solid black line represents the gradual
  shift of the attack point $x_{c}^{(p)}$ toward a local maximum.  The hinge
  loss and the classification error are shown in colors, to appreciate
  that the hinge loss provides a good approximation of the
  classification error. The value of such functions for each point $x
  \in [-5,5]^{2}$ is computed by learning an SVM on $\mathcal{D}_{\rm tr} \cup
  \{x,y=-1\}$ and evaluating its performance on $\mathcal{D}_{\rm
    val}$. The SVM solution on the clean data $\mathcal{D}_{\rm tr}$,
  and the training data itself, are reported for completeness,
  highlighting the support vectors (with black circles), the 
  decision hyperplane and the  margin bounds (with black lines).}
\label{fig:exp-artificial}
\end{center}
\end{figure*}

In the experiment presented below, the red class is the \emph{attacking}
class. To this end, a random point of the blue class is selected
and its label is flipped to serve as the starting point for our method. 
Our gradient ascent method is then used to refine this attack
until its termination condition is satisfied. The attack's trajectory is traced as the black line
in
Fig.~\ref{fig:exp-artificial} for both the linear kernel (upper two plots)
and the RBF kernel (lower two plots). The background in each plot
represents the error surface explicitly computed for all points within the
box $x \in [-5,5]^{2}$. The leftmost plots in each pair show the
hinge loss computed on a validation set while the rightmost plots in each pair
show the classification error for the area of interest. For the
linear kernel, the range of attack points is limited to the box
$x \in [-4,4]^{2}$ shown as a dashed line.   

For both kernels, these plots show that our gradient ascent
algorithm finds a reasonably good local maximum of the
non-convex error surface. For the linear kernel, it terminates at the
corner of the bounded region, since the error surface is
unbounded. For the RBF kernel, it also finds a good local maximum of the
hinge loss which, incidentally, is the maximum classification error within
this area of interest. 

\subsection{Real data}

We now quantitatively validate the effectiveness of the proposed
attack strategy on a well-known MNIST handwritten digit classification
task \cite{LeCun95}. Similarly to \citet{GloRow06}, we
focus on two-class sub-problems of discriminating between two distinct
digits.\footnote{The data set is also publicly available in Matlab
  format at \url{http://cs.nyu.edu/~roweis/data.html}.}  In
particular, we consider the following two-class problems: 7 vs. 1; 9 vs.
8; 4 vs. 0. The visual nature of the handwritten digit data provides us with
a \emph{semantic meaning} for an attack.  

Each digit in the MNIST data set is properly normalized and
represented as a grayscale image of $28 \times 28$ pixels. In particular, each pixel is ordered in a raster-scan and its value is directly considered as a feature. The overall number of features is $d = 28 \times 28 = 784$. We normalized each feature (pixel value) $x \in [0,1]^{d}$ by dividing its value by $255$.

In this experiment only the linear kernel is considered, and the
regularization parameter of the SVM is fixed to $C=1$. We randomly
sample a training and a validation data of $100$ and
$500$ samples, respectively, and retain the complete testing data given by MNIST
for $\mathcal D_{\rm ts}$. Although it varies for each digit,
the size of the testing data is about 2000 samples per class (digit).

\begin{figure*}[tb]
\begin{center}
\includegraphics[width=0.7\textwidth]{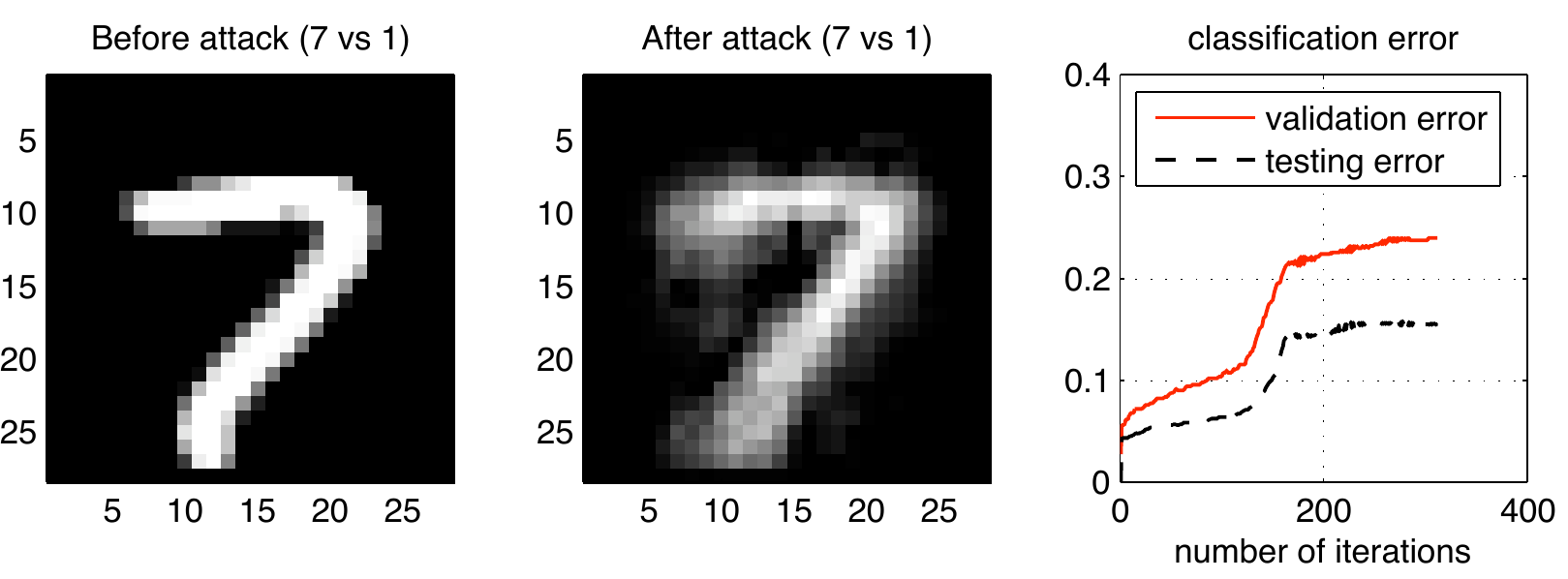}\\
\includegraphics[width=0.7\textwidth]{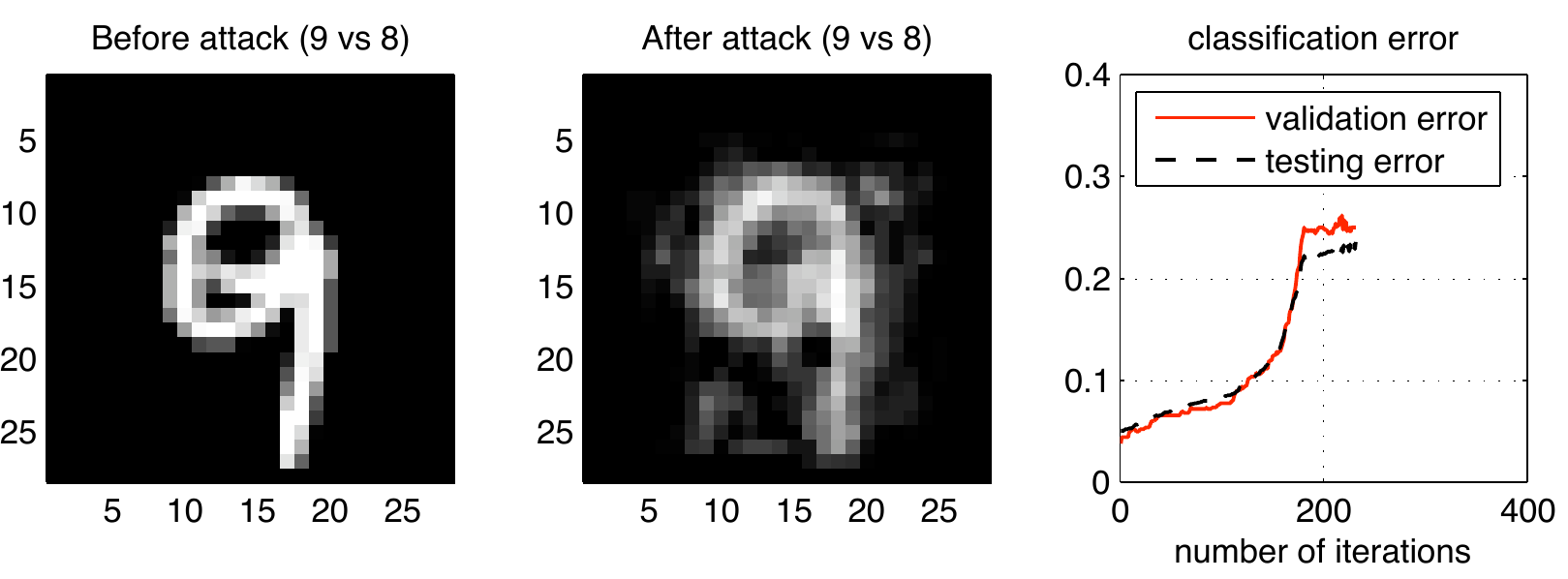}\\
\includegraphics[width=0.7\textwidth]{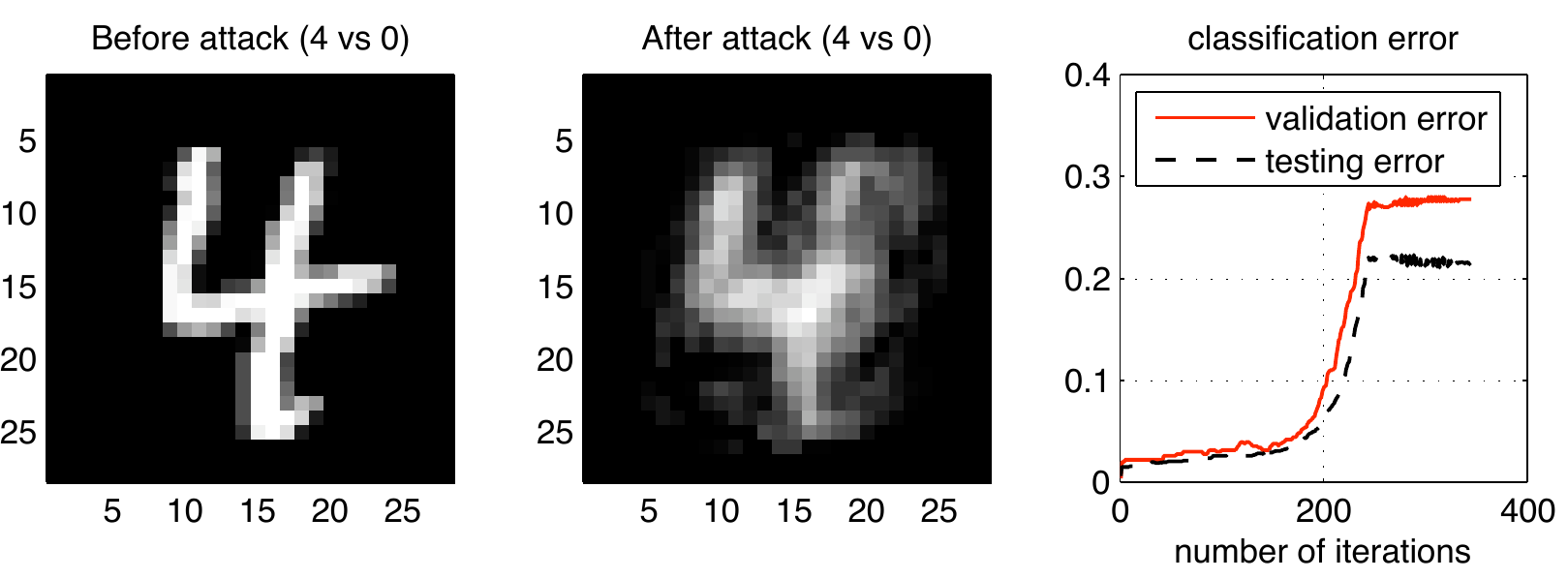}
\caption{Modifications to the initial (mislabeled) attack point
  performed by the proposed attack strategy, for the three considered
  two-class problems from the MNIST data set. The increase in
  validation and testing errors across different iterations is also
  reported.}
\label{fig:exp-single-digits}
\end{center}
\end{figure*}

The results of the experiment are presented in
Fig.~\ref{fig:exp-single-digits}. The leftmost plots of each row show the
example of the attacked class taken as starting points in our
algorithm. The middle plots show the final attack point. The rightmost
plots displays the increase in the validation and testing errors as
the attack progresses. 

The visual appearance of the attack point reveals that the attack
blurs the initial prototype toward the appearance of examples of the
attacking class. Comparing the initial and final attack points, we see this effect: the bottom segment of the $7$ straightens to resemble a
$1$, the lower segment of the $9$ becomes more round thus mimicking
an $8$, and \emph{round} noise is added to the outer boundary
of the $4$ to make it similar to a $0$.

The increase in error over the course of
attack is especially striking, as shown in the rightmost plots. In
general, the validation error overestimates the classification error
due to a smaller sample size. Nonetheless, in the exemplary runs
reported in this experiment, a single attack data point caused the
classification error to rise from the initial error rates
of 2--5\% to 15--20\%.
%BAT: comment on point 3
Since our initial attack point is obtained by flipping the label of a point in the attacked class, the errors in the first iteration of the rightmost plots of Fig.~\ref{fig:exp-single-digits} are caused by single random label flips. 
%
%This finding underscores the vulnerability of the SVM to poisoning attacks. 
This confirms that our attack can achieve significantly higher error rates than random label flips, and underscores the vulnerability of the SVM to poisoning attacks.

\begin{figure}[tbhp]
\begin{center}
\includegraphics[width=0.38\textwidth]{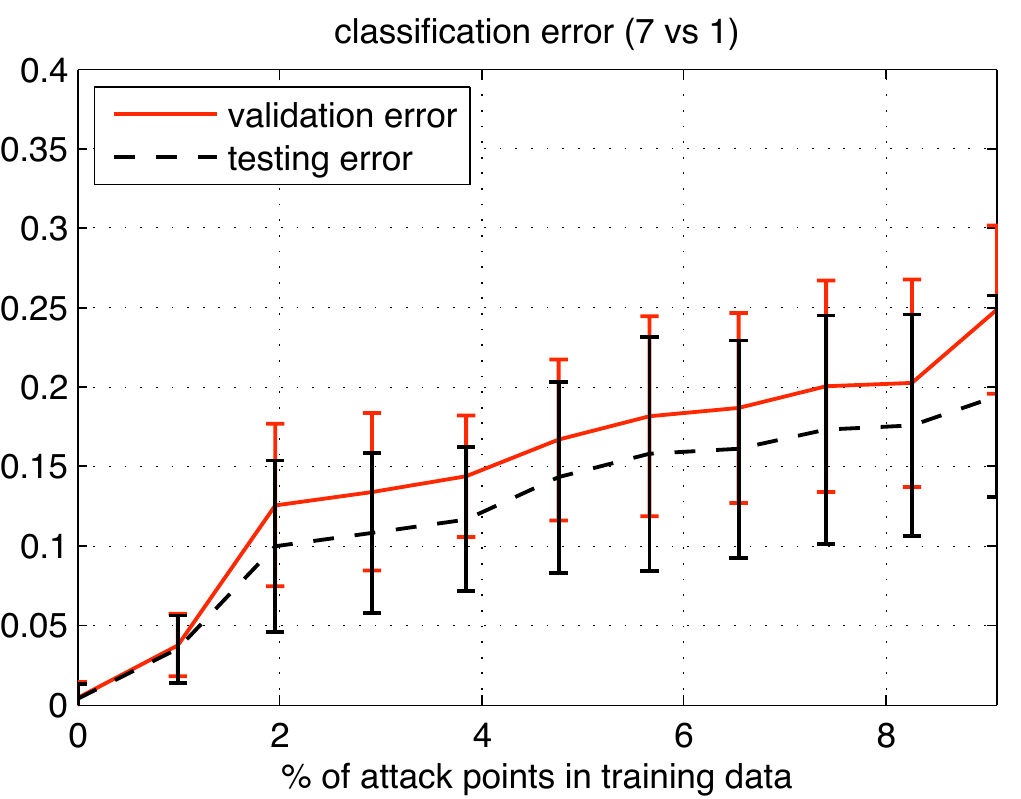}\\
\vspace{2em}
\includegraphics[width=0.38\textwidth]{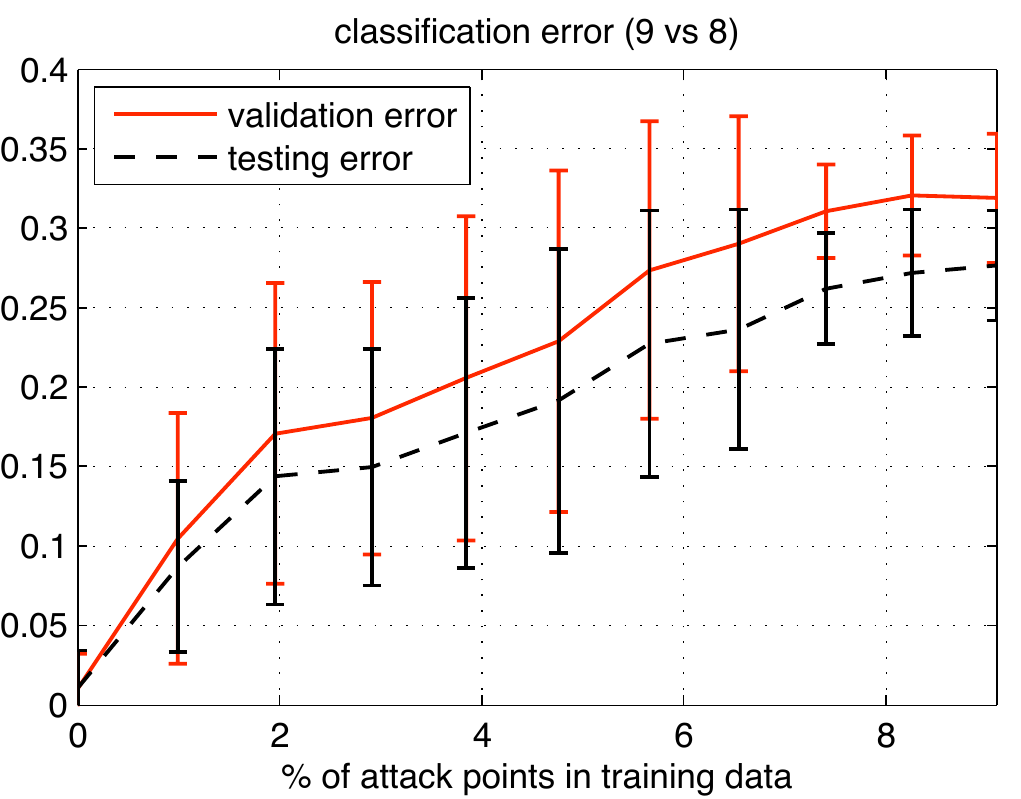}\\
\vspace{2em}
\includegraphics[width=0.38\textwidth]{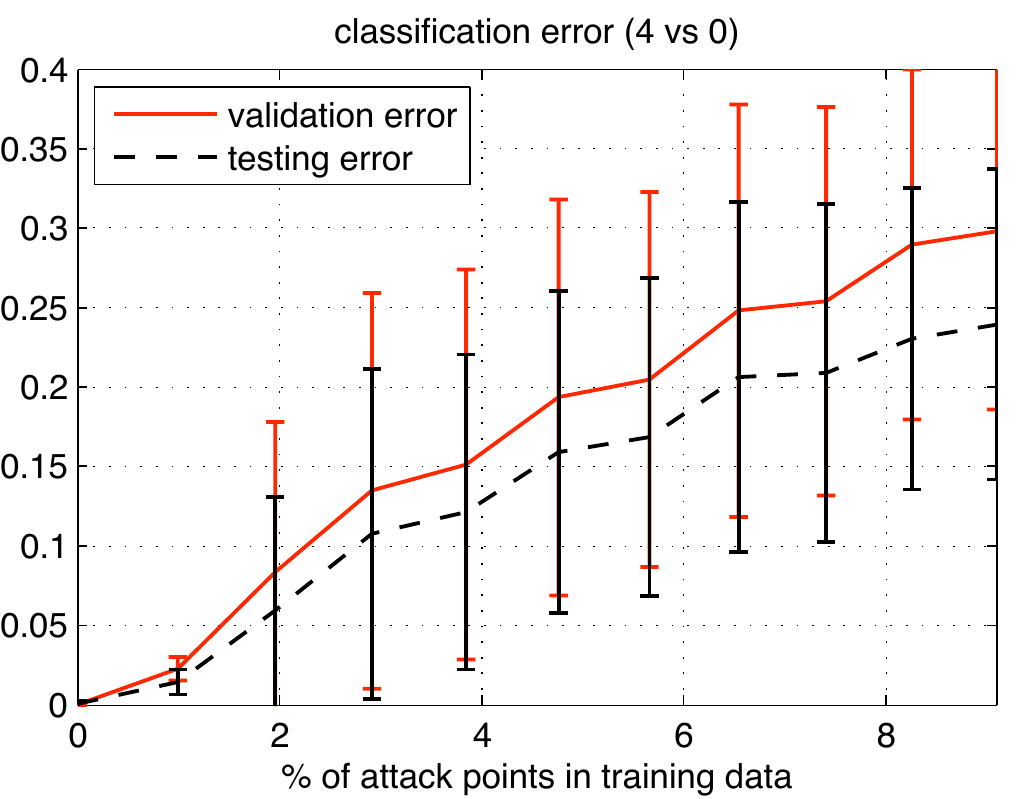}
\caption{Results of the multi-point, multi-run experiments on the
  MNIST data set. In each plot, we show the classification errors due to poisoning as a function of the percentage of training contamination for both the validation (red solid line) and testing sets (black dashed line). The topmost plot is for the $7$ vs.$1$ classifier, the middle is for the $9$ vs. $8$ classifier, and the bottommost is for the $4$ vs. $0$ classifier.}
\label{fig:exp-multi-digits}
\end{center}
\end{figure}

The latter point is further illustrated in a multiple point, multiple
run experiment presented in Fig.~\ref{fig:exp-multi-digits}. For this
experiment, the attack was extended by injecting additional points
into the same class and averaging results over multiple runs on
randomly chosen training and validation sets of the same size (100 and
500 samples, respectively). One can clearly see a steady growth of the
attack effectiveness with the increasing percentage of the attack
points in the training set. The variance of the error is quite high,
which can be explained by relatively small sizes of the training and
validation data sets. 

\section{Conclusions and Future Work}

% Optimal Steps: Using the work Pavel suggested on QCQPs to determine an optimal step to the edge of the structural changes.

The poisoning attack presented in this paper is the first step toward
the security analysis of SVM against training data attacks. Although
our gradient ascent method is arguably a crude algorithmic
procedure, it attains a surprisingly large impact on the
SVM's empirical classification accuracy. The presented attack method also reveals the
possibility for assessing the impact of transformations carried out in
the input space on the functions defined in the Reproducing Kernel
Hilbert Spaces by means of differential operators. Compared to
previous work on evasion of learning algorithms
\citep[\emph{e.g.},][]{Bruckner-Scheffer-NIPS-2009,Kloft-Laskov-AIStats-2010},
such influence may facilitate the practical realization of various
evasion strategies.  These implications need to be
further investigated.

Several potential improvements to the presented method remain to
be explored in future work.  The first would be
to address our optimization method's restriction to small changes in
order to maintain the SVM's structural constraints.  We solved this by
taking many tiny gradient steps. It would be interesting to
investigate a more accurate and
efficient computation of the largest possible step that does not alter
the structure of the optimal solution.

% Non-error Perturbations: How to perturb support vectors.

Another direction for research is the simultaneous optimization of multi-point
attacks, which we successfully approached with sequential single-point
attacks. The first question is how to optimally perturb a subset of
the training data; that is, instead of individually optimizing each
attack point, one could derive simultaneous steps for every attack
point to better optimize their overall effect. The second question is
how to choose the best subset of points to use as a starting point for
the attack. Generally, the latter is a subset selection
problem but heuristics may allow for improved approximations.
Regardless, we demonstrate that even non-optimal
multi-point attack strategies significantly degrade the
SVM's performance.

% Subterfuge: How can the attacker best mask his attack; perhaps by
% bounding $g_c$ (the margin of the attack point) within the context
% of the optimization over the test data.

An important practical limitation of the proposed method is the
assumption that the attacker controls the labels of the injected
points. Such assumptions may not hold when the labels
are only assigned by trusted sources such as humans. For instance,
a spam filter uses its users'
labeling of messages as its ground truth. Thus, although an attacker can
send arbitrary messages, he cannot guarantee that they will have the
labels necessary for his attack. This imposes an additional requirement
that the attack data must satisfy certain side constraints to
fool the labeling oracle. Further work is needed to understand these
potential side constraints and to incorporate them into attacks. 

The final extension would be to incorporate the real-world inverse
feature-mapping problem; that is, the problem of finding real-world
attack data that can achieve the desired result in the learner's input
space. For data like handwritten digits, there is a direct mapping
between the real-world image data and the input features used for
learning. In many other problems (\emph{e.g.}, spam filtering) the mapping is
more complex and may involve various non-smooth operations and
normalizations. Solving these inverse
mapping problems for attacks against learning remains open.

\section*{Acknowledgments}

This work was supported by a grant awarded to B.~Biggio by Regione
Autonoma della Sardegna, and by the project No. CRP-18293 funded by the same institution, PO Sardegna FSE 2007-2013, L.R.~7/2007 ``Promotion of the scientific research and technological innovation in Sardinia''. 
The authors also wish to acknowledge the Alexander von
Humboldt Foundation and the Heisenberg Fellowship of the Deut\-sche
Forschungsgemeinschaft (DFG) for providing financial support to carry
out this research. The opinions expressed in this paper are solely
those of the authors and do not necessarily reflect the opinions of
any sponsor.

{\small
\bibliography{sources}
\bibliographystyle{icml2012}
}
\end{document}